\documentclass{article}
\usepackage{spconf,amsmath,graphicx}
\usepackage{booktabs}
\usepackage{xcolor}
\usepackage{amssymb}

\title{Beyond Weak Labels: Prompt-Guided Local Refinement for Weakly Supervised Water Segmentation in High-Resolution Multispectral Imagery}

\name{
\begin{tabular}{c}
Muhammad Farhan Humayun$^{1}$ \quad Mohammad Imangholiloo$^{2}$ \quad Afifah Shah$^{1}$\\
Tomi Westerlund$^{1}$ \quad Jukka Heikkonen$^{1}$
\end{tabular}
}

\address{
$^{1}$Department of Computing, University of Turku, Finland\\
$^{2}$Department of Geoinformatics and Cartography, Finnish Geospatial Research Institute,\\
National Land Survey of Finland, Espoo, Finland
}

\begin{document}
%
\maketitle
\begin{abstract}

High-resolution water mapping supports environmental monitoring and related applications, but accurate pixel-level labels are difficult and costly to produce. Official hydrographic vectors provide scalable weak supervision, but they contain artifacts like boundary noise, temporal mismatch, and omissions of small water structures. We propose a two-stage framework for weakly supervised water segmentation in high-resolution multispectral imagery. Stage~1 learns initial masks from rasterized vector pseudo-labels, and Stage~2 converts these masks into structured component-wise prompts for localized refinement. On a manually corrected validation set, refinement improves SegFormer-B0 from 0.9509 to 0.9535 IoU and U-Net from 0.9408 to 0.9486 IoU, with corresponding F1 gains from 0.9749 to 0.9762 and 0.9695 to 0.9736. It leads to sharper shorelines, reduced boundary spillover, and better thin-structure delineation. The results indicate that prompt-guided refinement can improve pseudo-label-based water segmentation by targeting local errors that are poorly captured by global training supervision.

\end{abstract}

\section{Introduction}
\label{sec:intro}


Accurate water-body delineation from high-resolution multispectral imagery is important for hydrologic monitoring, flood response, environmental assessment, and geospatial inventory maintenance~\cite{LI2022306,WIELAND2023113452}. Compared with coarse-resolution products, high-resolution orthophotos derived from aerial and satellite imagery better capture object-level features such as shorelines, narrow channels, and small inland water bodies~\cite{essd-14-3349-2022, isprs-annals-X-2-W2-2025-65-2025}. However, they also introduce substantial challenges such as water boundaries are irregular, target structures are often thin and fragmented, and visually similar non-water regions can cause local ambiguities~\cite{Cao2024,9883460}. A major difficulty in this setting is the lack of reliable pixel-level ground truth. Hydrographic mapping vector layers are attractive because they are widely available and inexpensive compared with exhaustive manual annotation, but they are not perfect for supervision in deep learning models as they often contain boundary noise, temporal mismatch with the imagery, geometric simplification, and omission of small water bodies or narrow channels. They are therefore useful as pseudo-labels, but not as exact ground truth, making weak supervision a natural formulation for this problem~\cite{9668924, 9992041}.  

Under these constraints, modern segmentation architectures such as SegFormer have shown strong segmentation performance and robustness across image resolutions~\cite{NEURIPS2021_64f1f27b}. Such models are well suited as strong baselines for weakly supervised remote sensing segmentation because they can learn from large pseudo-labeled datasets while retaining the capacity to model both local detail and broader scene context~\cite{9759447}. Yet high aggregate accuracy alone is not sufficient in this setting. In high-resolution water mapping, many tiles are dominated by large easy regions of land or open water, and global pixel-wise scores can therefore hide crucial errors that matter most in practice including shoreline placement, thin channels, small omitted water bodies and  islands~\cite{Cheng_2021_CVPR}. The central challenge is thus not only coarse water detection, but the refinement of fine-scale local structure under imperfect supervision.

\begin{figure*}[t]
    \centering
    \includegraphics[width=16cm]{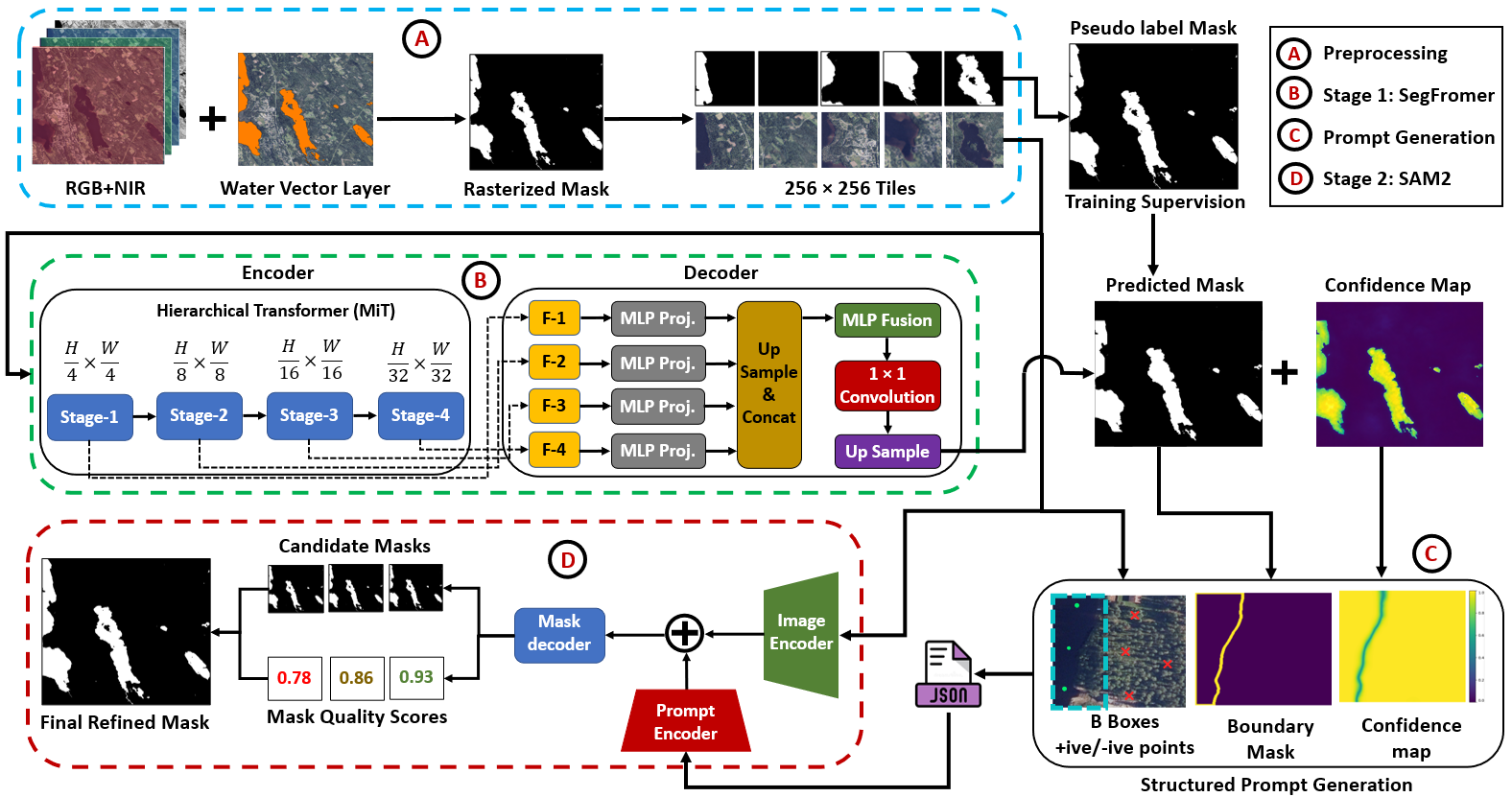}
    \caption{Prompt-guided two-stage pipeline: (A) preprocessing, (B) weakly supervised SegFormer segmentation, (C) structured prompt generation, and (D) localized SAM 2 mask refinement.}
    \label{full-pipeline}
\end{figure*}

This research proposes a two-stage framework for weakly supervised water segmentation. The first stage produces an initial water mask using pseudo-label supervision, while the second stage performs \emph{prompt-guided local refinement} using the Segment Anything Model 2 (SAM 2)~\cite{ravi2025sam2}. In this formulation, the refinement stage is guided by explicit local instructions that indicate where refinement should occur and how the promptable model should interpret each candidate region. The resulting two-stage formulation is conceptually simple and practically useful. Our contributions are threefold: (1) a weakly supervised multispectral water-segmentation pipeline; (2) a structured component-wise, prompt-generation scheme for localized mask refinement; and (3) evidence that local prompt-guided refinement improves a strong baseline, particularly for boundary adjustment and recovery of small water/land components.

\section{Materials and Methods}
\label{sec:format}

The study area is located in Joensuu, Finland, and includes diverse water bodies such as lakes, ponds, river channels, and narrow streams. We use high-resolution orthophotos from National Land Survey of Finland (NLS), combining RGB and NIR bands into four-channel multispectral inputs. Weak supervision is obtained by rasterizing hydrographic vector layers from the Finnish Environment Institute (SYKE) into pseudo-label masks.

\subsection{Pre-processing and Pseudo-labels}
Figure~\ref{full-pipeline} summarizes the two-stage pipeline. SYKE hydrographic vector layers were rasterized to the 50 cm orthophoto grid and used as weak pseudo-labels. The four-channel images and masks were tiled into $256 \times 256$ patches. All water-containing tiles were retained, including sparse-water and boundary cases, while pure-land tiles were heavily subsampled as hard negatives to emphasize shorelines, water-land transitions, and narrow structures. Since the pseudo-labels contained boundary noise, temporal mismatch, simplification artifacts, and omissions, evaluation used a manually corrected strong-label subset from held-out regions, rasterized to the same grid, extent, resolution, and coordinate reference system. The final dataset contained $8184$ training tiles and $810$ strong-label validation tiles, approximately a $90{:}10$ split.


\subsection{Weakly Supervised Baseline Segmentation}

Stage 1 generates the initial water prediction for each multispectral tile. Let 
$\mathbf{X}_i \in \mathbb{R}^{H \times W \times 4}$ denote the $i$-th RGB-NIR input tile and 
$\tilde{\mathbf{Y}}_i \in \{0,1\}^{H \times W}$ denote its rasterized pseudo-label mask. We use SegFormer-B0 as the primary baseline because it provides a favorable balance between efficiency and segmentation performance. The model is initialized with pretrained weights and adapted to binary water segmentation using four-channel inputs. The network predicts pixel-wise logits:
\begin{equation}
\mathbf{Z}_i = f_{\theta}(\mathbf{X}_i), \qquad 
\mathbf{Z}_i \in \mathbb{R}^{H \times W \times 2},
\end{equation}
where $f_{\theta}$ denotes the SegFormer network and the two output channels correspond to non-water and water, respectively. Pixel-wise class probabilities are then obtained by a softmax operation:
\begin{equation}
\mathbf{P}_i = \mathrm{softmax}(\mathbf{Z}_i), \qquad
\mathbf{P}_i \in [0,1]^{H \times W \times 2}.
\end{equation}
Training is performed using pseudo-label masks derived from official vector layers, whereas evaluation is conducted only on manually corrected strong labels. The network is optimized using pixel-wise cross-entropy loss for binary semantic segmentation:
\begin{equation}
\mathcal{L}_{\mathrm{seg}} =
-\sum_{u,v} \sum_{c=1}^{2}
\tilde{Y}_{i,c}(u,v)\log P_{i,c}(u,v).
\end{equation}
The final binary prediction is obtained by selecting the most probable class at each pixel:

\begin{equation}
\hat{\mathbf{Y}}_i(u,v) = \arg\max_{c \in \{0,1\}} P_{i,c}(u,v).
\end{equation}
To support Stage 2 refinement, we also derive a confidence map from the water-class probability. Let $P_{i,1}(u,v)$ denote the predicted probability of the water class. The confidence is defined as the distance from the decision boundary:
\begin{equation}
\mathbf{C}_i(u,v) = 2\left| P_{i,1}(u,v) - 0.5 \right|.
\end{equation}
Thus, pixels with values close to $0.5$ are treated as uncertain, while pixels close to $0$ or $1$ are treated as confident predictions. The Stage-1 output consists of the binary mask $\hat{\mathbf{Y}}_i$ and confidence map $\mathbf{C}_i$, which together provide the coarse water extent and uncertainty cues for prompt-guided local refinement in Stage 2. We use SegFormer-B0 as the primary Stage-1 baseline because it provides a favorable balance between efficiency and segmentation performance. For comparison, we also experimented with a classical U-Net backbone~\cite{10.1007/978-3-319-24574-4_28} as an alternative Stage-1 baseline under the same weakly supervised training and evaluation protocol.

\subsection{Structured Spatial Prompt Generation}

This module provides the bridge between Stage 1 and Stage 2. For each Stage-1 prediction, connected water components are extracted from the binary mask and converted into structured spatial prompts. Let $\hat{\mathbf{Y}}_i \in \{0,1\}^{H \times W}$ denote the Stage-1 binary prediction for tile $i$, and let $\mathbf{C}_i \in [0,1]^{H \times W}$ denote the corresponding confidence map. The set of connected water components is defined as
\begin{equation}
\mathcal{K}_i = \mathrm{CC}(\hat{\mathbf{Y}}_i),
\end{equation}
where $\mathrm{CC}(\cdot)$ denotes connected-component extraction. For each connected component $k \in \mathcal{K}_i$, its spatial support is defined as
\begin{equation}
\Omega_{ik} = \{(u,v)\mid (u,v)\ \text{belongs to component } k\}.
\end{equation}
The corresponding bounding box is then computed as
\begin{equation}
\mathbf{b}_{ik} = [x_{\min}, y_{\min}, x_{\max}, y_{\max}],
\end{equation}
Each component is represented by a structured prompt object
\begin{equation}
\pi_{ik} =
\{\mathbf{b}_{ik}, \mathcal{P}^{+}_{ik}, \mathcal{P}^{-}_{ik}, r_{ik}, m_{ik}\},
\end{equation}
where $\mathbf{b}_{ik}$ is the component bounding box, $\mathcal{P}^{+}_{ik}$ and $\mathcal{P}^{-}_{ik}$ denote sets of positive and negative point prompts, $r_{ik}$ denotes the local refinement region, and $m_{ik}$ denotes the refinement mode. Positive points are sampled from the component interior and high-confidence foreground support, while negative points are sampled from nearby non-water, boundary-adjacent, or hole-like regions depending on the local geometry and confidence cues.

The refinement mode specifies how Stage 2 should interpret the local region, for example whether refinement is focused on boundary correction, hole preservation, or suspicious-component recovery. These structured prompt objects are stored in per-tile JSON files for reproducibility and inspection. During Stage-2 inference, the JSON contents are parsed and converted into SAM 2-compatible spatial inputs.






\subsection{Prompt-Guided Local Mask Refinement}

Stage 2 refines the Stage-1 prediction using SAM 2 in a local, component-wise manner. For each tile, the RGB image $\mathbf{I}_i^{\mathrm{RGB}}$ and the structured prompt object $\pi_{ik}$ are provided to SAM 2 to generate one or more candidate masks for component $k$:
\begin{equation}
\mathcal{M}_{ik} = g_{\phi}(\mathbf{I}_i^{\mathrm{RGB}}, \pi_{ik}),
\end{equation}
where $g_{\phi}$ denotes the promptable segmentation model and $\mathcal{M}_{ik}$ is the set of candidate masks produced for the local region.

The generated candidates are not accepted directly. Instead, each candidate mask $\mathbf{M}_{ik}^{j} \in \mathcal{M}_{ik}$ is filtered using geometric consistency checks with respect to the Stage-1 component and its allowed refinement region. In particular, we require sufficient agreement with the reference component and constrain the relative area change:
\begin{equation}
\frac{|\mathbf{M}_{ik}^{j} \cap \Omega_{ik}|}{|\Omega_{ik}|} \geq \alpha,
\qquad
\beta_{\min} \leq
\frac{|\mathbf{M}_{ik}^{j}|}{|\Omega_{ik}|}
\leq \beta_{\max},
\end{equation}
where $\Omega_{ik}$ is the spatial support of the Stage-1 component, $\alpha$ is the minimum overlap threshold, and $\beta_{\min}$ and $\beta_{\max}$ define acceptable lower and upper bounds on area change. Additional mode-dependent screening is applied to prevent implausible updates outside the allowed local region.


\begin{table*}[t]
\centering
\caption{Performance comparison of weakly supervised Stage-1 baselines and their corresponding SAM 2-based prompt-guided refinements on the manually corrected strong-label validation set.}
\setlength{\tabcolsep}{5pt}
\renewcommand{\arraystretch}{1.08}
\footnotesize
\begin{tabular}{cccccccccc}
\toprule
\textbf{Method} & \textbf{TP (\%)} & \textbf{FP (\%)} & \textbf{FN (\%)} & \textbf{TN (\%)} & \textbf{IoU} & \textbf{F1} & \textbf{Precision} & \textbf{Recall} & \textbf{Accuracy} \\
\midrule
Stage 1 (SegFormer-B0) & 41.40 & 1.42 & 0.72 & 56.46 & 0.9509 & 0.9748 & 0.9669 & \textbf{0.9829} & 0.9786 \\
Stage 2 (SAM 2 Refinement) & 41.34 & \textbf{1.24} & 0.78 & \textbf{56.65} & \textbf{0.9535} & \textbf{0.9762} & \textbf{0.9709} & 0.9814 & \textbf{0.9798} \\
\midrule
Stage 1 (U-Net) & 40.71 & 1.16 & 1.41 & 56.72 & 0.9408 & 0.9695 & 0.9723 & \textbf{0.9666} & 0.9743 \\
Stage 2 (SAM 2 Refinement) & \textbf{40.97} & \textbf{1.07} & \textbf{1.15} & \textbf{56.81} & \textbf{0.9485} & \textbf{0.9736} & \textbf{0.9745} & \textbf{0.9726} & \textbf{0.9778} \\
\bottomrule
\label{tab-results}
\end{tabular}
\end{table*}

After filtering, the selected candidate $\mathbf{M}_{ik}^{*}$ is merged back into the Stage-1 prediction only within a restricted local refinement region $r_{ik}$:
\begin{equation}
\hat{\mathbf{Y}}^{\,\mathrm{ref},k}_i(u,v) =
\begin{cases}
\mathbf{M}_{ik}^{*}(u,v), & (u,v) \in r_{ik},\\
\hat{\mathbf{Y}}_i(u,v), & (u,v) \notin r_{ik},
\end{cases}
\end{equation}
and the final refined mask is obtained by applying this update sequentially over the component set. Thus, regions outside the allowed refinement zone preserve the original Stage-1 prediction. This prevents the promptable model from unnecessarily rewriting already plausible large water bodies. The refinement branch therefore acts primarily as a boundary-aware local correction module i.e., it targets uncertain boundaries, small structures, and locally ambiguous regions while preserving the coarse semantic layout learned by Stage 1.

\subsection{Implementation Details}

Stage~1 used the SegFormer-B0 model initialized from the official pretrained checkpoint. The architecture was kept unchanged except for a four-channel RGB--NIR input layer and a binary water/non-water output head. The model was fine-tuned for 50 epochs with batch size 32, learning rate $6 \times 10^{-5}$, and weight decay $1 \times 10^{-4}$. Stage~2 used the official SAM~2 image predictor with the SAM~2.1 Hiera-small checkpoint. SAM~2 was not modified; refinement was controlled externally using JSON-derived spatial prompts, geometric candidate filtering, and local mask-merging rules. All experiments were run in Python/PyTorch under Anaconda on an NVIDIA RTX 3500 Ada GPU with 12~GB VRAM and 32~GB RAM. Performance was evaluated on the manually corrected strong-label validation set using pooled IoU, F1-score, precision, recall, overall accuracy, and pixel-level TP, FP, TN, and FN counts.



\vspace {0.05cm}

\section{Results and Discussion}

Table~\ref{tab-results} compares Stage-1 weakly supervised baselines with their SAM~2-based prompt-guided refinements on the manually corrected validation set. For SegFormer-B0, Stage~2 reduced false positives by 12.8\% (685,840 to 598,124), improving precision from 0.9669 to 0.9710, accuracy from 0.9786 to 0.9799, IoU from 0.9509 to 0.9535, and F1-score from 0.9749 to 0.9762. Recall slightly decreased from 0.9829 to 0.9815, indicating that refinement mainly suppressed false water detections while introducing a small increase in missed water pixels.

For U-Net, SAM~2 refinement reduced both false positives and false negatives respectively. This improved IoU from 0.9408 to 0.9486, F1-score from 0.9695 to 0.9736, recall from 0.9666 to 0.9727, and accuracy from 0.9744 to 0.9778. The consistent gains across both Transformer-based and convolutional baselines indicate that the structured prompt-guided branch improves weakly supervised water segmentation beyond a single backbone.

Although the aggregate gains are modest, they are meaningful for high-resolution water mapping, where large easy land/open-water regions can hide local errors in global metrics. Stage~2 is therefore best interpreted as a boundary-aware local refinement module rather than a semantic replacement for Stage~1. Its main effect is improved shoreline alignment and correction of difficult local structures, with occasional recovery of severely incorrect Stage-1 regions.

\begin{figure}[htb]

  \centering
  \centerline{\includegraphics[width=6cm]{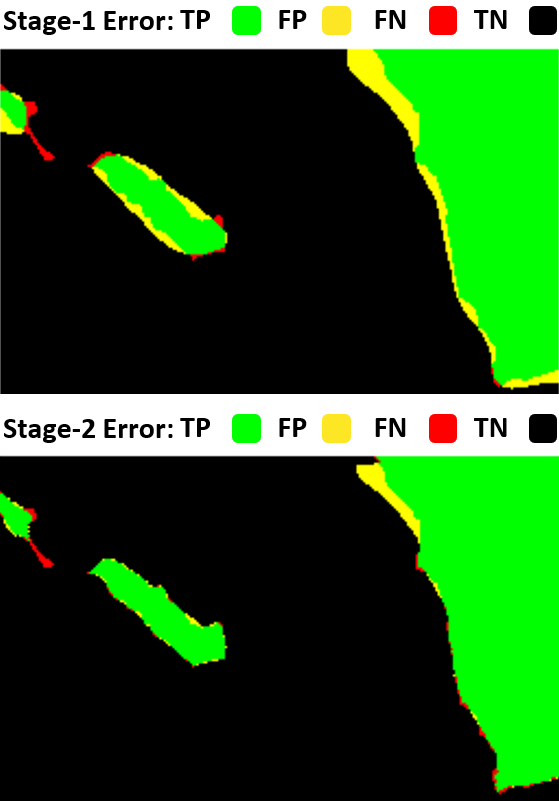}}

\caption{Zoomed-in local error maps before and after refinement of SegFormer-B0 visualizing improvement of water-body delineations.}
\label{fig:zoomed}
\end{figure}

\begin{figure*}[!t]
    \centering
    \includegraphics[width=12.5cm]{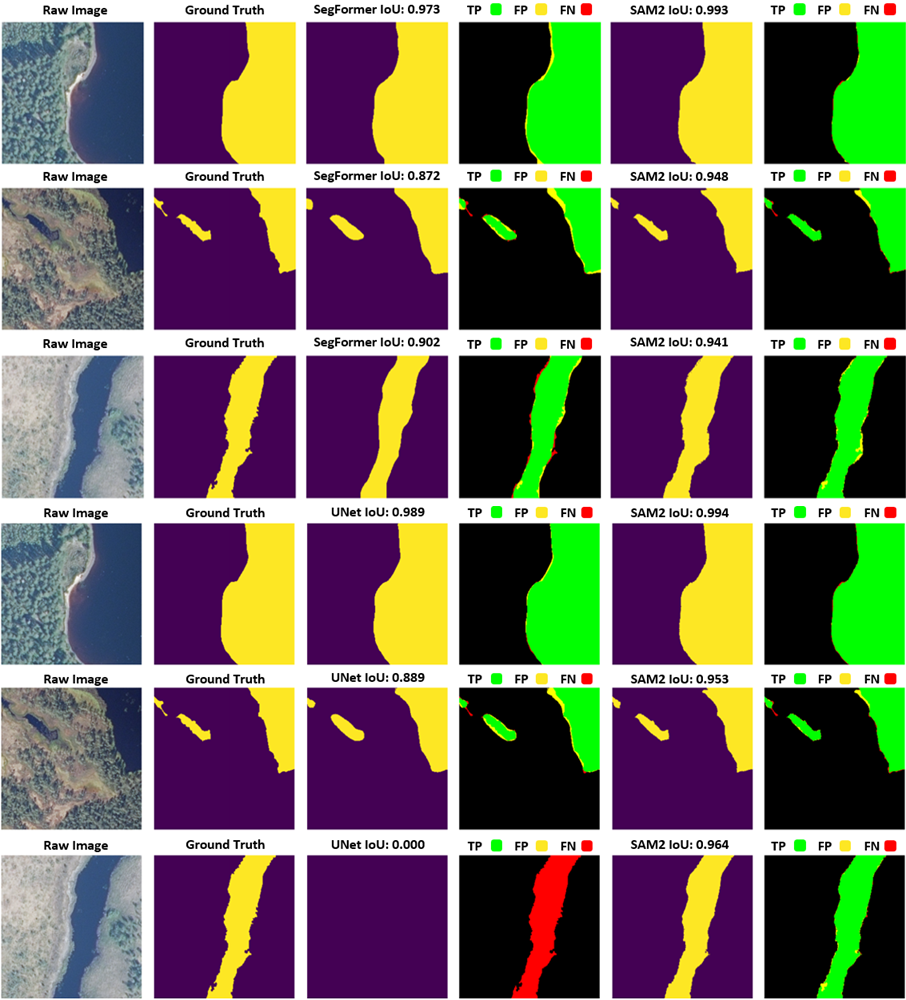}
    \caption{Visual Comparison of stage 1, weak baseline segmentation and stage 2 prompt-based refinement. Top three rows correspond to SegFormer-B0 as the stage 1 model whereas bottom three rows show results from U-Net as stage 1 model.}
    \label{fig:visual-results}
\end{figure*}

The visual results confirm that Stage~2 mainly performs local boundary correction rather than global mask rewriting. In Fig.~\ref{fig:zoomed}, refinement reduces shoreline spillover and produces cleaner water-land transitions while preserving the main Stage-1 water structure. Fig.~\ref{fig:visual-results} further shows the improved local mask quality for both SegFormer-B0 and U-Net, especially along irregular shorelines, narrow channels, and fragmented water structures. For SegFormer-B0, tile-level IoU improves in representative cases from 0.872 to 0.948 and from 0.902 to 0.941, mainly through reduced boundary spillover. For U-Net, refinement similarly improves boundary placement and suppresses local false positives. In one difficult case, the baseline fails almost completely, while Stage~2 recovers a plausible water mask and improves IoU from 0.000 to 0.964. These examples are consistent with Table~\ref{tab-results} and support Stage~2 as a boundary-aware local correction module whose success still depends on Stage-1 mask quality and generated prompts.

\section{Conclusion}
\label{sec:foot}

This study presented a two-stage weakly supervised framework for high-resolution multispectral water segmentation. Stage~1 used SegFormer-B0 and U-Net to learn from rasterized official hydrographic pseudo-labels, while Stage~2 converted their predictions into structured spatial prompts for localized SAM~2 refinement, evaluated on a manually corrected strong-label subset. SAM~2 refinement improved both baselines: SegFormer-B0 increased from 0.9509 to 0.9535 IoU and U-Net from 0.9408 to 0.9486 IoU, with corresponding F1 gains from 0.9749 to 0.9762 and 0.9695 to 0.9736. In representative tiles, the proposed approach produced larger local gains in tile-level IoU by reducing false spillover in uncertain boundary regions. The gains were mainly due to reduced false positives, sharper shoreline placement, and better handling of narrow or fragmented water structures. The results show that prompt-guided refinement is effective as a boundary-aware local correction module. This provides a practical route for improving pseudo-label-based water segmentation under limited annotations, especially for fragmented or narrow water bodies with shoreline spillover. Future work will explore richer terrain and spectral cues and test transferability across high-resolution Earth observation settings.

\vspace {-0.1cm}
\section{ACKNOWLEDGEMENTS}
\label{sec:copyright}

This research has been conducted with Flagship Programme funding granted by the Research Council of Finland for Digital Waters Flagship (decision no. 359247 and decision no. 359249). This research was also supported by the Postdoctoral Programme for Research Institutes in Finland funded by the Finnish Government.


\bibliographystyle{IEEEbib}
\bibliography{strings,refs}

\end{document}